\documentclass[letterpaper, 10 pt, conference]{ieeeconf}  

\IEEEoverridecommandlockouts                              

\usepackage{xspace,empheq,fancybox,amsmath,amssymb,graphicx,epstopdf,epsfig,syntonly} 
\usepackage{bm}
\usepackage{url,cite,footnote,xspace,syntonly}
\usepackage{verbatim,multirow}
\usepackage[T1]{fontenc}
\usepackage{mathtools}
\usepackage{mwe}
\usepackage{algorithm}
\usepackage{algpseudocode}
\usepackage{graphicx}
\usepackage{subcaption}
\usepackage{textcomp}
\usepackage{xcolor}
\usepackage{multicol}
\usepackage{siunitx,booktabs}
\usepackage{csvsimple}
\usepackage{setspace}
\allowdisplaybreaks

\begin{document}
\title{\LARGE \bf Deep Reinforcement Learning-Based Approach for a Single Vehicle Persistent Surveillance Problem with Fuel Constraints}
\author{Manav Mishra$^\dag$, Hritik Bana$^\dag$, Saswata Sarkar$^\dag$, Sujeevraja Sanjeevi$^\ddag$, PB Sujit$^\dag$, and Kaarthik Sundar$^*$ 
\thanks{
$^\dag$Multi-Robot Autonomy Laboratory, IISER, Bhopal, India. \newline E-mail: \texttt{\{hritik19, mishra20, saswata19, sujit\}@iiserb.ac.in}}
\thanks{
$^\ddag$OpsLab Inc., Austin, Texas, USA.\@ \newline  E-mail: \texttt{sujeev.sanjeevi@gmail.com}}
\thanks{
$^*$Los Alamos National Laboratory, New Mexico, USA.\@ \newline E-mail: \texttt{kaarthik@lanl.gov}
}}

\maketitle
\thispagestyle{empty}
\pagestyle{empty}

\begin{abstract}
This article presents a deep reinforcement learning-based approach to tackle a persistent surveillance mission requiring a single unmanned aerial vehicle initially stationed at a depot with fuel or time-of-flight constraints to repeatedly visit a set of targets with equal priority. Owing to the vehicle's fuel or time-of-flight constraints, the vehicle must be regularly refueled, or its battery must be recharged at the depot. The objective of the problem is to determine an optimal sequence of visits to the targets that minimizes the maximum time elapsed between successive visits to any target while ensuring that the vehicle never runs out of fuel or charge. We present a deep reinforcement learning algorithm to solve this problem and present the results of numerical experiments that corroborate the effectiveness of this approach in comparison with common-sense greedy heuristics.
\end{abstract}


\section{Introduction} \label{sec:intro}
Continuous observation and analysis of diverse environments ranging from infrastructure networks to remote military bases traditionally involve manual, repetitive, and time-sensitive tasks. Over the past decade, the advent of Unmanned Aerial Vehicles (UAVs), with their ease of deployment \cite{austin2011unmanned}, has made them ideal candidates for use in the aforementioned continuous observation and analysis problems. A Persistent Surveillance Mission (PSM) is a modeling paradigm used to schedule tasks for the vehicles in such problems that are repetitive and last for prolonged periods. In this work, we consider a PSM with a single UAV with a time-of-flight restriction tasked with surveilling a set of distinct targets; it is also assumed that every target in the target set has the same priority. The UAV is initially stationed at a depot and serves as a refueling or battery swap station that the UAV can periodically use. The objective of the PSM is to determine a sequence of visits to the targets and the depot that minimizes the maximum time elapsed between successive visits to any target, referred to as the maximum revisit time of the target. Minimizing the maximum revisit time over the target set will ensure that the vehicle surveils all targets as frequently as possible, and periodic depot visits ensure that the UAV never runs out of fuel or battery charge as it surveils the target locations. Throughout the rest of the article, we refer to the above PSM problem as the Single Vehicle Persistent Surveillance with Fuel Constraints (SVPSFC).

The choice of Reinforcement Learning (RL) as an algorithmic approach for the SVPSFC is motivated by two key factors. Firstly, though, one can formulate the SVPSFC as a mixed-integer linear program by leveraging existing work for fuel-constrained vehicle routing problems and the PSMs in \cite{hari2020optimal,hari2022bounds,sundar2013algorithms,sundar2016formulations}, these formulations are notoriously hard to solve to optimality for state-of-the-art branch-and-cut approaches. Even for problem instances that can be optimally solved, the time taken by these algorithms to solve the problem instances is prohibitively large \cite{hari2020optimal}. Hence, the standard approach to tackle the SVPSFC is to either relax the fuel restrictions of the vehicle or replace the fuel restrictions using surrogates like restricting the number of target visits before returning to the depot \cite{hari2020optimal}. These relaxations or surrogates make the problem easier to solve at the cost of sub-optimality and practical implementability of the solutions. The second factor is the design of either heuristics or approximation algorithms \cite{alamdari2014persistent} is also non-trivial and variant-specific, i.e., they will work when fuel constraints are ignored and do not directly extend to accommodate the fuel restrictions on the vehicle. Over the past couple of years, RL has proved to be successful in solving numerous mission planning problems with one or more UAVs \cite{yue2022deep,yuksek2021cooperative,liu2022hybrid,maw2021iada}, and given that traditional approaches have difficulties in addressing the SVPSFC, this paper formulates the SVPSFC as a Markov Decision Process (MDP) that Deep-RL (D-RL) can solve.

\section{Related Work} \label{sec:lit-work}
The problem of persistent surveillance was initially introduced by authors in \cite{las2013persistent}, and it varies from other exploratory missions in the sense that it requires repeated visitation of targets. Since its inception, a multitude of problem variants have garnered considerable attention from researchers (see \cite{hari2020optimal} and references therein). One can consider the PSM problem as a persistent area coverage or routing problem. In persistent area coverage, the goal is to persistently monitor areas in a given environment using a fleet of mobile guards with a fixed visibility range (see \cite{o1987art, tokekar2015visibility, maini2018, mishra2021multi}). A common approach to achieve the said area coverage is to employ a cellular decomposition that partitions the area into cells and assigns one or more agents to each cell for monitoring the area \cite{nigam2008persistent, elmaliach2009multi}. Here, the utilization of the visibility range constructs a visibility graph that, in turn, models constraints on the path taken by an agent enforced by obstacles. Extensive research has directed towards modeling the PSM as a routing problem. In particular, it is modeled as a multiple watchman routing problem \cite{yu2014, wang2010generalized, Hari2019, stump2011multi, alamdari2014persistent}. The only difference between the watchman routing problem and the set-up considered in this article is that, unlike UAVs with fuel or time of flight restrictions, the watchman routing problem imposes no such restrictions on the watchmen. These approaches aim to determine routes for each watchman to monitor a set of locations while minimizing visit latency. Here, latency for a location is measured as the maximum weighted or unweighted revisit time, with weights used to model priorities. Solution approaches to solve the multiple watchmen routing problem range from heuristics to approximation algorithms due to the computational complexity of the PSM \cite{alamdari2014persistent}.

One drawback with the existing approaches to modeling the PSM as an area coverage or a routing problem is that none consider the vehicle's fuel constraints, i.e., they neglect the vehicle's fuel capacity. Incorporating fuel restrictions on the UAV through periodic refuel stops at the depot would align it well with existing UAV capabilities and lead to a straightforward transition to practice. This drawback has also been pointed out in the literature \cite{nigam2008persistent}. It is also known that accounting for the fuel restrictions is an essential part of the planning problem, and neglecting the same can lead to routes with notably higher revisit times in practice than those planned for \cite{nigam2008persistent}. At this juncture, we also remark surrogates for the fuel restrictions, namely a fixed number of target visits before the need for refueling, have been proposed recently for the PSM in \cite{Hari2019} and \cite{Hari2021}. While this is a reasonable surrogate, the work does not delve into how a practitioner can consistently compute a reliable estimate of this number. This work seeks to remedy the drawbacks discussed in this paragraph by directly modeling and tackling the fuel restrictions of the UAV in a D-RL framework. To the best of our knowledge, this is the first work in the literature that seeks to tackle the fuel restrictions of the UAV directly and solve the SVPSFC using the D-RL approach.  

The rest of the article is organized as follows: Sec. \ref{sec:formulation} presents notation later used to state the SVPSFC problem mathematically. This is followed by the methodology section in \ref{sec:methodology} that formulates the problem as an MDP and presents the main ingredients of the D-RL approach used to solve the problem. Finally, results of extensive computational experiments that corroborate the effectiveness of the proposed approach are presented in Sec. \ref{sec:results} followed by the conclusion and avenues for future work in Sec. \ref{sec:conclusion}.

\begin{figure}
    \centering
    \includegraphics[width=0.4\textwidth
    ]{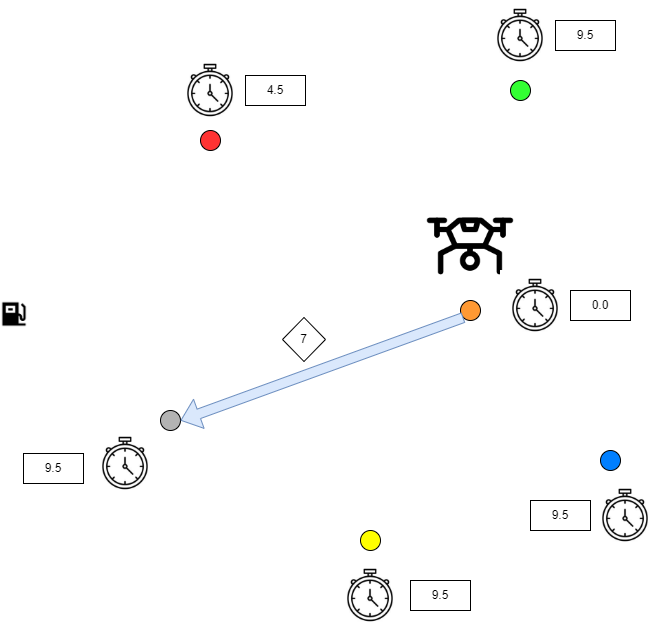}
    \caption{Environment with a UAV that visits the targets represented by the colored dots. Each target has an associated clock that shows the time elapsed since the last visit to that target. In this particular instance, the UAV is at the target colored in orange, and consequently, its clock reflects the time as zero. Once the vehicle reaches the target,  it decides to visit some other target (grey-colored targets), in this case, located at a distance of $7$ units from the target colored in orange.}
    \label{fig:problem-formulation}
\end{figure}

\section{Notations \& Problem Definition} \label{sec:formulation}
We start by introducing notations that will be used to formally define the SVPSFC.\@ The SVPSFC problem is formulated in a 2-dimensional environment. We note that this assumption is made for ease of explanation, and all the formulations and algorithms presented in this article also extend to a higher dimensional setting. The environment contains $n$ targets or points of interest and a depot at which a UAV is initially stationed. The set of $n$ targets and the depot is denoted by $\mathcal T \triangleq \{\tau_0, \tau_1, \dots, \tau_n\}$ with $\tau_0$ denoting the depot and $\tau_1, \dots, \tau_n$ denoting the $n$ targets. Given these notations, the SVPSFC is formulated on an undirected graph $\mathcal G \triangleq (\mathcal T, \mathcal E)$ where $\mathcal T$ denotes the vertex set and the set $\mathcal E \triangleq \{(i, j): i, j \in \mathcal T \text{ and } i \neq j\}$ denote the edge set. Without loss of generality, we also assume that the vehicle travels with a constant speed of $1$ unit. Hence, the travel time between any two vertices equals the travel distance between the pair. We assume that the travel distance between $\tau_i$ and $\tau_j$ is given by the Euclidean distance between them and is denoted by $d_{ij}$. Finally, we also assume that fuel consumed by the vehicle to traverse an edge $(i, j) \in \mathcal E$ is proportional to the distance between them and that the vehicle's fuel capacity is given by $F$.  For ease of exposition, we assume that the constant of proportionality is exactly $1.0$, making the fuel consumed to traverse the edge $(i, j)$ exactly equal to $d_{ij}$. Furthermore, we let $f$ denote the fuel left in the vehicle at any time during the mission. At the start of the mission, $f = F$, and whenever the vehicle traverses an edge $(i, j)$, $f$ is updated as $f = f - d_{ij}$. When the vehicle visits the depot $\tau_0$, it is assumed that it refuels to its full capacity, i.e., $f = F$.  To enable easy computation of the maximum revisit time for a target, we assume each target in $\tau_i \in \mathcal T\setminus \{\tau_0\}$ is associated with a clock $c_i$ that is used to track the time since the vehicle last visited $\tau_i$. Whenever the vehicle visits a target $\tau_i \in \mathcal T\setminus \{\tau_0\}$, its clock $c_i$ resets to zero. The clock update scheme when a vehicle reaches target $\tau_j$ from an arbitrary target $\tau_i$ is given by 
\begin{gather}
    c_k = \begin{cases}
        c_k + d_{ij} & \text{when $\tau_k \neq \tau_j$} \\ 
        0 & \text{when $\tau_k = \tau_j$}
    \end{cases}
    \label{eq:clock-update}
\end{gather}
The clock update scheme in \eqref{eq:clock-update} shows that the clock $c_i$ is reset to zero when the vehicle visits target $\tau_i$.

To track the revisit time of a target $\tau_i$, we introduce another variable, $r_i$. This variable updates when a target is visited more than once by the vehicle.  Now, suppose that the UAV visits the target $\tau_i$ from some arbirary target $\tau_j$, then $r_i$ is updated as follows:
\begin{gather}
    r_i = \begin{cases} 
    \max \{r_i, c_i + d_{ji}\} & \text{when number of $\tau_i$ visits $> 2$} \\ 
    c_i + d_{ji} & \text{when number of $\tau_i$ visits $= 2$} \\ 
    \infty &    \text{otherwise.}
    \end{cases} \label{eq:revisit_time}
\end{gather}
When a target $\tau_i$ is visited exactly twice and when the target that the vehicle visited right before $\tau_i$ is $\tau_j$, the revisit time takes meaningful values as dictated by \eqref{eq:revisit_time}. When $\tau_i$ is either not visited or visited precisely once by the UAV, we let $r_i$ take the value of $\infty$ as shown in \eqref{eq:revisit_time}. 

Though the set-up of the SVPSFC problem is such that it can generate an infinite sequence of target visits, we restrict our attention to $m\cdot n$ visits. The parameter $m$ can be changed to suit the duration for which persistent surveillance needs to be performed by the vehicle. Using all the above notations, the objective of the SVPSFC is mathematically formulated as 
\begin{gather}
    \min ~\{r_1, \dots, r_n\} \label{eq:obj}
\end{gather}
where $r_i$ is given by \eqref{eq:revisit_time}. This setup of the SVPSFC in this article imposes no priorities on the targets. If one chooses to model priorities, it is easy to introduce weights for each target and use them in the clock update equations for the respective target in \eqref{eq:clock-update}. Intuitively, this corresponds to letting the clocks for the high-priority targets run faster, leading to a higher $r_i$ value for those targets. Finally, an illustration of the set-up is shown pictorially in Fig. \ref{fig:problem-formulation}. In the next section, we formulate the SVPSFC as an MDP and present the main ingredients of the D-RL approach to solving the MDP.\@

\section{Methodology} \label{sec:methodology}
D-RL algorithms traditionally are used to solve problems formulated as MDPs. This section starts by formulating the SVPSFC problem as an MDP.\@



\subsection{MDP Formulation} \label{subsec:mdp} 
MDPs provide a framework to model decision-making problems in which single or multiple agents interact with an environment and the outcome of the problem is partly or fully under the control of the agent(s). MDPs are defined using a 4-tuple $(\mathcal S, \mathcal A, \mathcal P_a, \mathcal R_a)$ where $\mathcal S$ is a set of possible states that the agent(s) can be in, $\mathcal A$ is a set of possible actions that agents can take when they are at a given state, $\mathcal P_a$ is the transition probability function which provides the probability of transitioning from one state to another under a given action $a \in \mathcal A$ and finally, $\mathcal R$ is the reward function that evaluates the value of taking action $a \in \mathcal A$. Notice that though an MDP is generally set up to take into account uncertainty through probability transition matrices, it is equally well suited to handle decision-making problems where one knows the state transitions with certainty, as in the case of the SVPSFC problem; in this scenario, the underlying MDP is described using $(\mathcal S, \mathcal A, \mathcal R)$. We now define the state space $\mathcal S$, the action space $\mathcal A$, and the reward function $\mathcal R$ for the MDP corresponding to SVPSFC.\@

The state space $\mathcal S$ informally contains all the information of the environment and is defined mathematically as $\mathcal S  \triangleq \langle \ell, \bm c, \bm d, \bm x, f\rangle$. The individual symbols in $\mathcal S$ are defined as follows:
\begin{itemize}
    \item $\ell$ -- the index of the vertex that the agent is currently at, i.e., when $\ell = i$, it implies that the agent is currently at target $\tau_i$. $\ell$ can take any value from the set $\{0, 1, \dots, n\}$. 
    \item $\bm c$ -- a vector in $\mathbb R^n_{\geqslant 0}$ where each element corresponds to the clock value $c_i$ of target $\tau_i$.
    \item $\bm d$ -- a vector in $\mathbb R^{n+1}_{\geqslant 0}$. Specifically, the $i$-th element of $\bm d$ is the distance between the vertex $\tau_i$ and the vertex corresponding to the current location index $\ell$. 
    \item $\bm x$ -- vector of cartesian coordinates of the $(n+1)$ vertices in the graph $\mathcal G$
    \item $f$ -- fuel left in the UAV; $f$ can take a value in $[0, F]$. 
\end{itemize}

The SVPSFC problem focuses on identifying a sequence of targets without considering the intermediary time taken for the agent's transition between locations. Therefore, in each step, the agent gets the information from the environment through $\mathcal S$ and selects the next target to visit. Hence, the action space is defined as $\mathcal A \triangleq \{0, 1, \dots, n\}$. If the UAV chooses an action $a \in \mathcal A$, it has decided to transition to vertex $\tau_a$. Now, we define the state transition rule associated with our MDP.\@ Note that there is no uncertainty in transitioning the agent from one target to another, so if the agent's initial state is $s \in S$ and it chooses an action $a \in A$, then it arrives at the new state $s' = \langle \ell^{\prime}, \bm c^{\prime}, \bm d^{\prime}, \bm x, f^{\prime} \rangle$, where,
\begin{gather}
    \ell^{\prime} = a \\
    c_i^{\prime} = \begin{cases}
        c_i + d_a & \text{ for $i \neq a$} \\ 
        0 & \text{ for $i = a$}
    \end{cases}  \\ 
    f^{\prime} = \begin{cases}
        f - d_a & \text{ for $a \neq 0$} \\ 
        F & \text{ for $a = 0$}
    \end{cases}
    \label{eq:state-transition}
\end{gather}
and finally $\bm d^{\prime}$ is updated based on the action $a$ using the distance of each edge in the edge set $\mathcal E$. The final piece of the MDP is the reward function. To align with the objective function in \eqref{eq:obj} we formulate the reward function of the MDP as 
\begin{gather}
    \mathcal R \triangleq -\max ~\{c_1, \dots, c_n\} \label{eq:reward-fcn}
\end{gather}

In general, dynamic programming is the algorithm of choice to solve any MDP with a finite state space. In our case, the state space $\mathcal S$ is infinite-dimensional since all the values in $\langle \ell, \bm c, \bm d, \bm x, f\rangle$ are continuous real numbers with some of them being non-negative reals. In such cases, solving the MDP using dynamic programming is difficult. Moreover, classical RL approaches such as Q-learning or SARSA also run into issues like infeasibilities due to the continuous nature of state-space \cite{sutton2018reinforcement}. Hence, we use a D-RL approach to compute the optimal policy. This paper uses a standard on-policy Proximal Policy Optimization (PPO) method over other policy gradient methods for its ability to avoid extensive policy updates, enhance learning stability, high sample efficiency, and resistance to hyper-parameter tuning \cite{yu2022surprising}.

\subsection{Techniques to improve transferability} \label{subsec:dummy-targets}
The D-RL approach to solving the MDP formulated in the previous section traditionally can work only for a fixed number of targets, i.e., if the number of targets changes, then the PPO has to be trained again for the new value of $n$. To address this issue, we propose the use of the so-called dummy targets. This will enable training for a single value of $n$ and re-using the same policy for any number of targets less than equal to $n$. Specifically, we let $K$ number of dummy targets where $0 \leqslant K \leqslant (n - 2)$. These dummy targets are not required to be visited by the vehicle, and their clocks and maximum revisit times in \eqref{eq:clock-update} and \eqref{eq:revisit_time}, respectively, are always set to $0$. Hence, they will never impact the reward calculation in \eqref{eq:reward-fcn}. Now, the D-RL is trained for a fixed value of $n$ and different values of $K$ in its domain. Thus, we start with $(n - K)$ real and $K$ dummy targets, train the D-RL over multiple episodes on this setup, and reset observations and environment initialization after each episode. Subsequently, we reinitialize the environment by randomly sampling $k'$ real and $(n - k')$ dummy targets, with $(n - k')$ and $k'$ distributed randomly across the spatial domain. This iterative process enables the agent to learn near-optimal paths across any configuration with fewer than $n$ targets.

\subsection{Action masking to enforce fuel restrictions} \label{subsec:action-space-maksing}
The approach presented in the previous section enables training a D-RL network for $n$ targets and using the same network to obtain the optimal policy of any instance with less than $n$ targets. Another constraint that can create issues is the fuel restrictions of the UAV, i.e., the reward function presents susceptibility to adverse scenarios, such as fuel depletion mid-mission or continual selection of the current location as the subsequent target, thereby hindering clock updates and optimizing rewards. Traditionally, this issue is tackled through reward shaping to discourage adverse actions. However, reward shaping requires a high degree of fine-tuning with regard to the extent of the penalty. High penalties for fuel depletion may result in the agent remaining stationary, while coupling this with penalties for prolonged stationarity leads the agent to adopt a conservative oscillatory strategy within a subset of targets. Furthermore, when the space of infeasible actions is enormous, it becomes highly unlikely for the agent ever to encounter one of the scarce feasible actions during the exploration phase of D-RL.\@

To address this issue, we utilize the concept of varying the action space depending on the feasibility of the visitation of targets, termed action masking \cite{action_mask}. This approach restricts the agent from executing infeasible actions. For the SVPSFC problem, action masking is implemented at three levels. Firstly, at each step, the action associated with the current target location is masked to prevent repetitive visits of the UAV to the same target. Secondly, all targets designated as dummy targets for the current simulation setup are masked. Thirdly, we calculate the distance for the agent to travel directly to the depot after visiting each target. Leveraging the triangle inequality, we mask all actions corresponding to locations where the current fuel, $f$, is insufficient to reach the depot based on this calculated distance. This ensures that only the subset of targets from which a refueling trip to the depot is feasible can be chosen, thereby ensuring the feasibility of the solution. 

The algorithmic approach and its enhancements presented thus far are a complete framework to solve the SVPSFC problem using D-RL.\@ In the next section, we present the results of extensive computational experiments that corroborate the approach's effectiveness in solving the SVPSFC problem.

\section{Experiments and Analysis} \label{sec:results}
In this section, we empirically evaluate the performance of the D-RL approach to the SVPSFC problem, focusing on the following questions: (1) How does the D-RL perform with increasing targets compared to a heuristic baseline? (2) How does the D-RL compare with the baseline for vehicles with varying fuel capacity? (3) How do trajectories obtained using the D-RL and the baseline vary qualitatively?

\subsection{Experimental Setup}
 We employ OpenAI's Gym library to implement our environment. Our environment is a 2-dimensional square grid of size 10 units, with the depot located at the origin $(0, 0)$ and other target locations scattered throughout the grid. The UAV's fuel capacity is set at $F = 120$ units, with a fuel consumption rate of one unit per unit distance traveled. We also assume that the UAV travels at a speed of one unit distance per unit of time. We trained our model on $n=14$ targets using the methodology presented in Sec. \ref{subsec:dummy-targets} and \ref{subsec:action-space-maksing}, effectively learning all configurations on targets less than $n$. To train D-RL model using the action-masked PPO approach, we employ the Stable Baselines 3 library \cite{stable-baselines3}. Each configuration is trained for 5000 steps, after which we randomly adjust the number and positions of targets. We conduct training for a total of 150 million steps. The policy network of the action-masked PPO algorithm is implemented using a Multi-layer Perceptron Policy \cite{stable-baselines3}. For comparison, we use 100 configurations for each $n \in \{2, 3, \dots, 14\}$ and compare the average maximum revisit time for each $n$ over the $100$ runs.



\subsection{Results}
To show the efficacy of the D-RL approach, we also implement a `baseline' greedy algorithm.  In the greedy baseline heuristic, the UAV selects the feasible target with the current highest clock time at each decision step. Additionally, target feasibility is determined by the remaining fuel capacity. If the UAV lacks sufficient fuel for a round trip to a target and back to the depot, that target is deemed infeasible. The episode ends when no feasible targets remain for the UAV to visit, resulting in a mission failure.


\subsubsection{Comparison of D-RL and the greedy baseline for increasing $n$} 
\begin{figure}[h]
    \centering
    \includegraphics[width=0.5\textwidth]{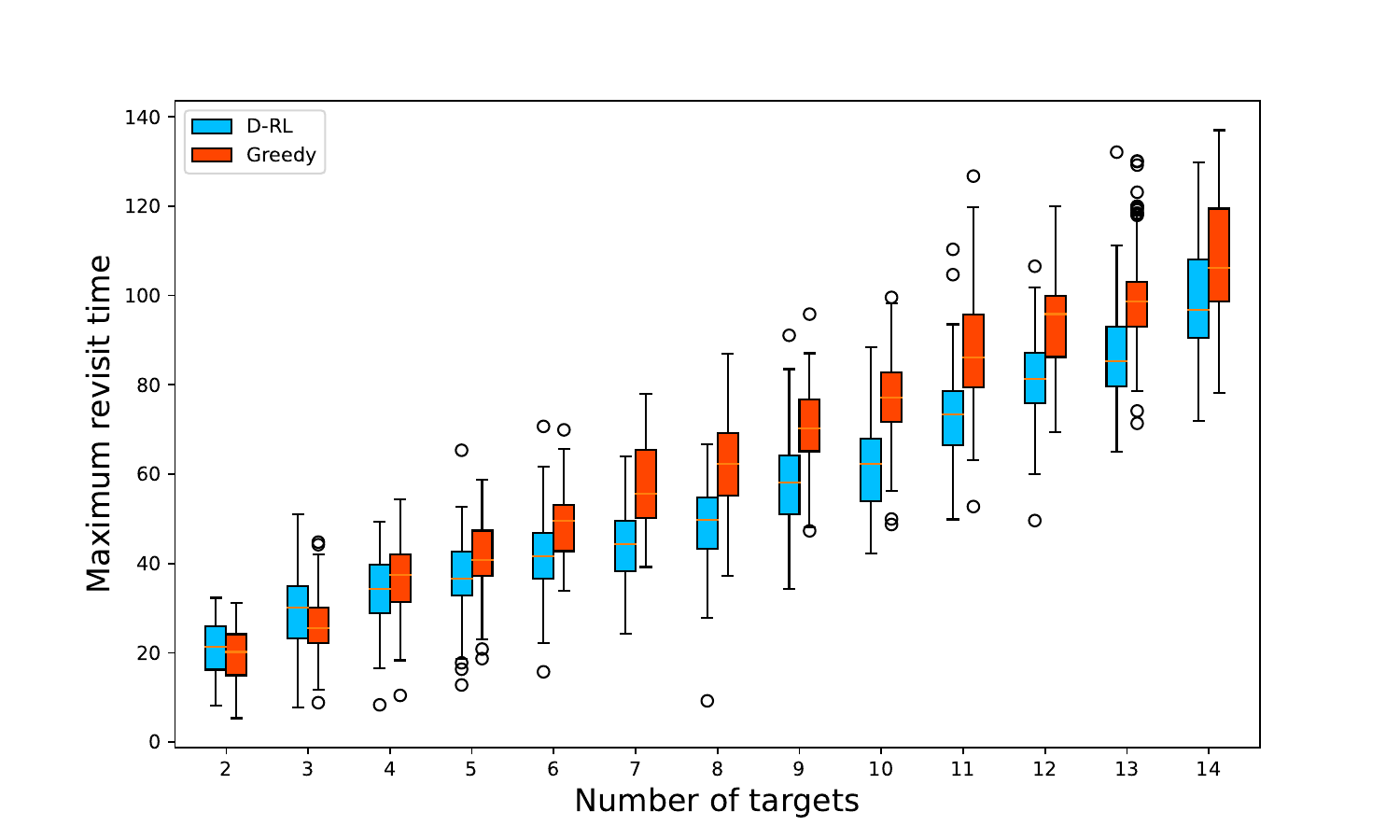} 
    \caption{Performance comparison of maximum revisit time for the D-RL and greedy approach. 
    }
\label{fig:results_14targ}
\end{figure}

Figure \ref{fig:results_14targ} shows the results for D-RL and the greedy baseline. The box plot shows the median maximum revisit time over 100 configurations for each $n \in \{2, 3, \dots, 14\}$. From Fig. \ref{fig:results_14targ}, it is evident that, except for the on-par result for $n = 2$, the D-RL approach consistently achieves a lower median maximum revisit time compared to the greedy baseline heuristic $\forall n > 2$. Notably, the D-RL and greedy approaches attain $100\%$ mission success, preventing any adverse occurrences such as fuel depletion; this was possible due to action-masking presented in Sec. \ref{subsec:action-space-maksing}.

\subsubsection{Comparison of D-RL and the greedy baseline for varying fuel capacity}
\begin{figure}
    \centering
\includegraphics[width=0.5\textwidth]{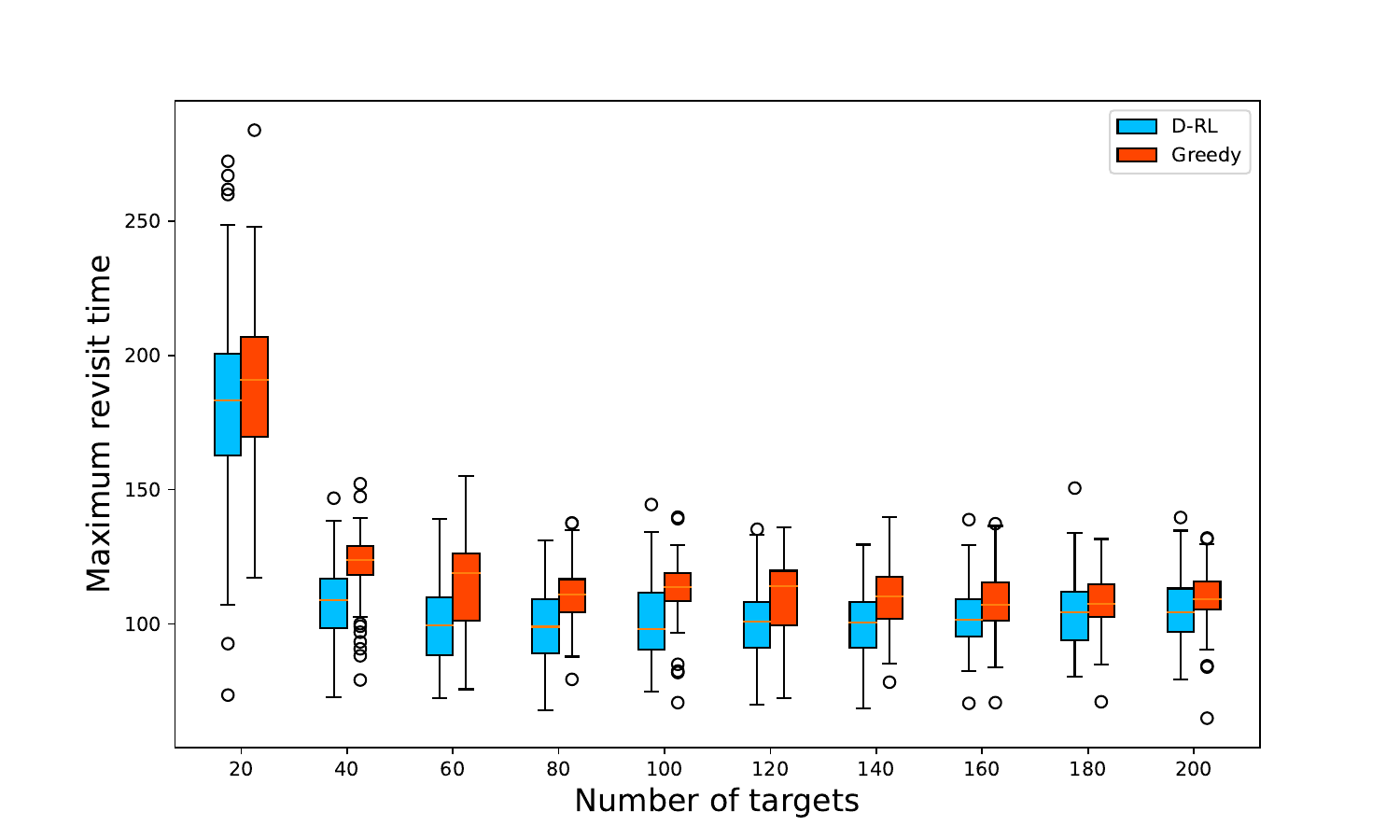}
    \caption{
    The average maximum revisit time for 100 test configurations is $n = 14$, while we vary the maximum fuel capacity. 
    }
    \label{fig:Time vs fuel}
\end{figure}

The D-RL agent shows robustness to variations in the fuel capacity of the UAV, eliminating the need for re-training when fuel capacity changes. As shown in Fig. \ref{fig:Time vs fuel}, the D-RL agent, initially trained with a maximum fuel capacity of $120$ units, maintains efficacy across a range of capacities ($\in \{20, 40, 60, 80, 100, 120, 140, 160, 180, 200\}$ units). 
The D-RL approach still consistently achieves a lower maximum revisit time for all the fuel capacities than the greedy baseline approach. For smaller fuel capacity, the average maximum revisit time is higher for both approaches because the lower capacity requires the UAV to repeatedly visit the depot for refueling, which in turn leads to higher revisit times for the targets. This highlights the D-RL algorithm's capacity to optimize resource utilization without requiring extensive reconfiguration of the model.

\subsubsection{Qualitative comparison of the trajectories} 
\begin{figure*}
\centering
\subfloat[]{\label{fig:env-config}\includegraphics[width=0.33\textwidth]{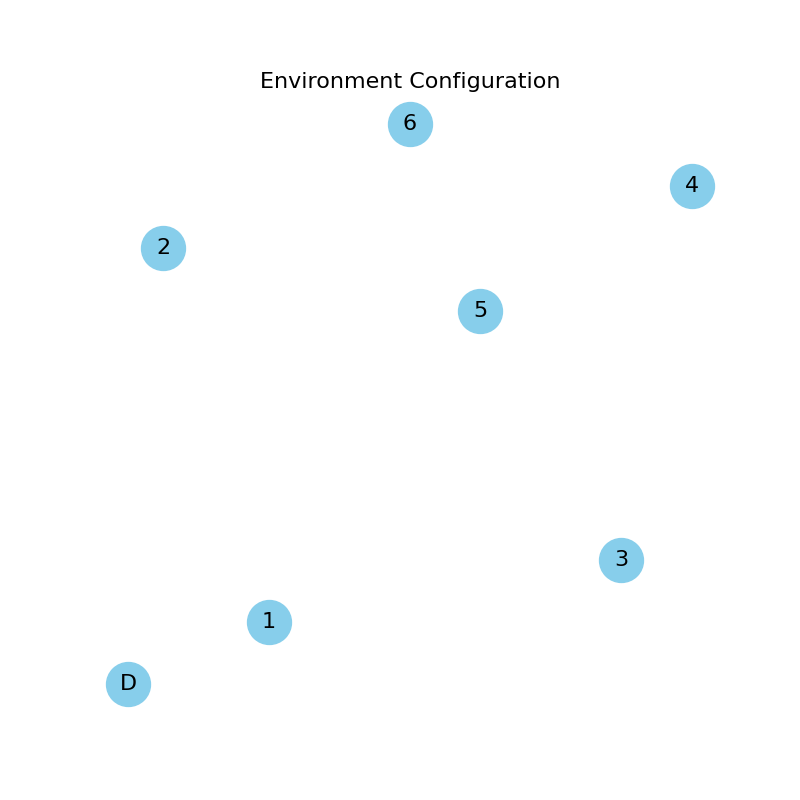}}
\subfloat[]{\label{fig:greedy-config}\includegraphics[width=0.33\textwidth]{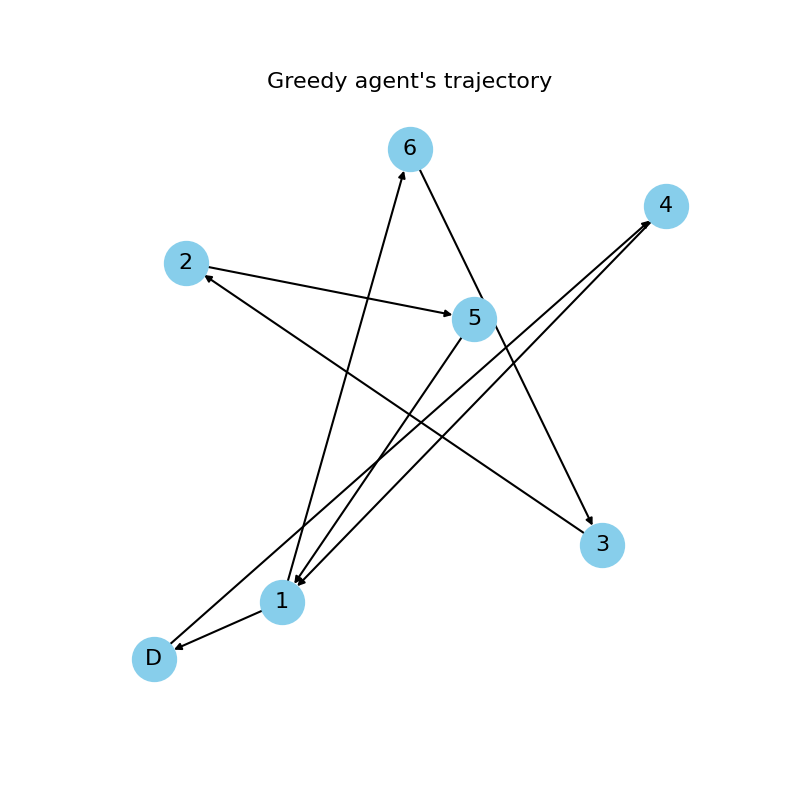}}
\subfloat[]{\label{fig:drl-config}\includegraphics[width=0.33\textwidth]{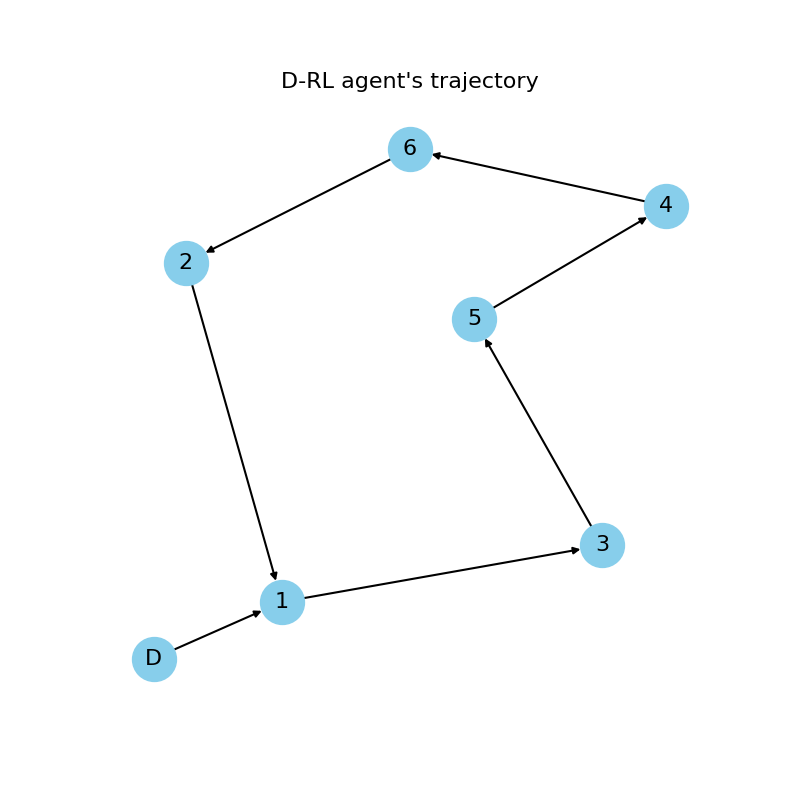}}
\caption{(a) Outline of the environment configuration for a set of six targets located at $\{(0, 0), (2, 1), (0.5, 7), (7, 2), (8, 8), (5, 6), (4, 9)\}$. (b) The first 8 steps of the UAV's trajectory obtained using the greedy approach. (c) The first 8 steps of the UAV's trajectory obtained using the D-RL approach.}\label{fig:qualitative-trajectory}
\end{figure*}

We analyze the trajectories from D-RL and greedy baseline approaches on a fixed configuration to examine both approaches' qualitative behaviors. Fig. \ref{fig:env-config} depicts these trajectories for $n = 6$ with fixed targets at coordinates $\{(0, 0), (2, 1), (0.5, 7), (7, 2), (8, 8), (5, 6), (4, 9)\}$. In the outlined configuration, Fig. \ref{fig:greedy-config} illustrates the first eight steps of the greedy approach's trajectory as $\{4, 1, 6, 3, 2, 5, 1, 0\}$. On running the entire mission for the specified $n \times m = 42$ steps, the maximum revisit time is $59.78$ units. Fig. \ref{fig:drl-config} shows the D-RL approach's first eight steps as $\{1, 3, 5, 4, 6, 2, 1, 3\}$. On running the entire mission for the specified $n \times m = 42$ steps, the maximum revisit time is $31.93$ units. The trajectory of the greedy approach reveals a tendency to visit distant targets early on, driven by the greedy approach favoring higher distances despite the resulting longer clock times. Consequently, the greedy approach frequently returns to the depot for refuelling within these initial steps, leading to higher maximum revisit times. In contrast, the D-RL trajectory demonstrates a more reasonable approach, preserving fuel reserves and reducing the need for frequent depot revisits, thus enabling a more efficient tour.

\section{Conclusion \& Future Work} \label{sec:conclusion}
In this article, we propose a D-RL-based solution approach for the Single Vehicle Persistent Surveillance with Fuel Constraint problem. Our primary contribution lies in addressing this problem without relying heavily on simplifications, thereby maintaining its applicability to real-world scenarios. The D-RL approach outperforms the baseline approach without encountering adverse situations, underscoring its utility in safety-critical contexts. We conduct a thorough analysis and compare our approach with existing methods, revealing its superiority in both performance and robustness. A natural extension of this study involves addressing targets with heterogeneous priorities, wherein certain specific targets must be visited comparatively more frequently than others. Another interesting study would be extending it to a multi-vehicle setting, where multiple agents have to perform the surveillance cooperatively. Optimal strategies would involve cooperative coordination to prevent redundant monitoring of the same targets by multiple vehicles and to minimize temporal trajectory overlap for efficient persistent surveillance.

\bibliography{references}

\bibliographystyle{IEEEtran}
\end{document}